\documentclass[10pt,twocolumn,letterpaper]{article}

\usepackage{cvpr}              % To produce the CAMERA-READY version
\definecolor{cvprblue}{rgb}{0.21,0.49,0.74}
\usepackage[pagebackref,breaklinks,colorlinks,allcolors=cvprblue]{hyperref}

\usepackage[table]{xcolor}
\usepackage{tikz}

\usepackage{multicol, multirow}

\def\paperID{*****} % *** Enter the Paper ID here
\def\confName{CVPR}
\def\confYear{2026}

\newcommand{\method}{Thermal-Det}

\title{\method: Language-Guided Cross-Modal \\Distillation for Open-Vocabulary Thermal Object Detection}

\author{Yasiru Ranasinghe\textsuperscript{1}, Elim Schenck\textsuperscript{2}, Florence Yellin\textsuperscript{2},Shuowen Hu\textsuperscript{3},\\Christopher Funk\textsuperscript{2}, and Vishal M. Patel\textsuperscript{1}\\
\textsuperscript{1}Johns Hopkins University $\cdot$ \textsuperscript{2}Kitware $\cdot$ \textsuperscript{3}DEVCOM Army Research Laboratory\\
{\small\ttfamily \{dranasi1, vpatel36\}@jhu.edu, shuowen.hu.civ@army.mil}\\ {\small\ttfamily \{christopher.funk, elim.schenck,florence.yellin\}@kitware.com,}
}

\begin{document}

% \AddToShipoutPicture*{%
%   \AtPageLowerLeft{%
%     \put(\LenToUnit{0.5in},\LenToUnit{0.35in}){%
%       \parbox{3in}{\small\ttfamily
%         \{dranasi1, vpatel36\}@jhu.edu\\
%         christopher.funk@kitware.com
%       }%
%     }%
%   }%
% }

\twocolumn[{%
  \maketitle
  \vspace{-0.75em} % tighten gap under title/ID; tweak or remove
  \begin{center}
    \includegraphics[width=0.99\linewidth]{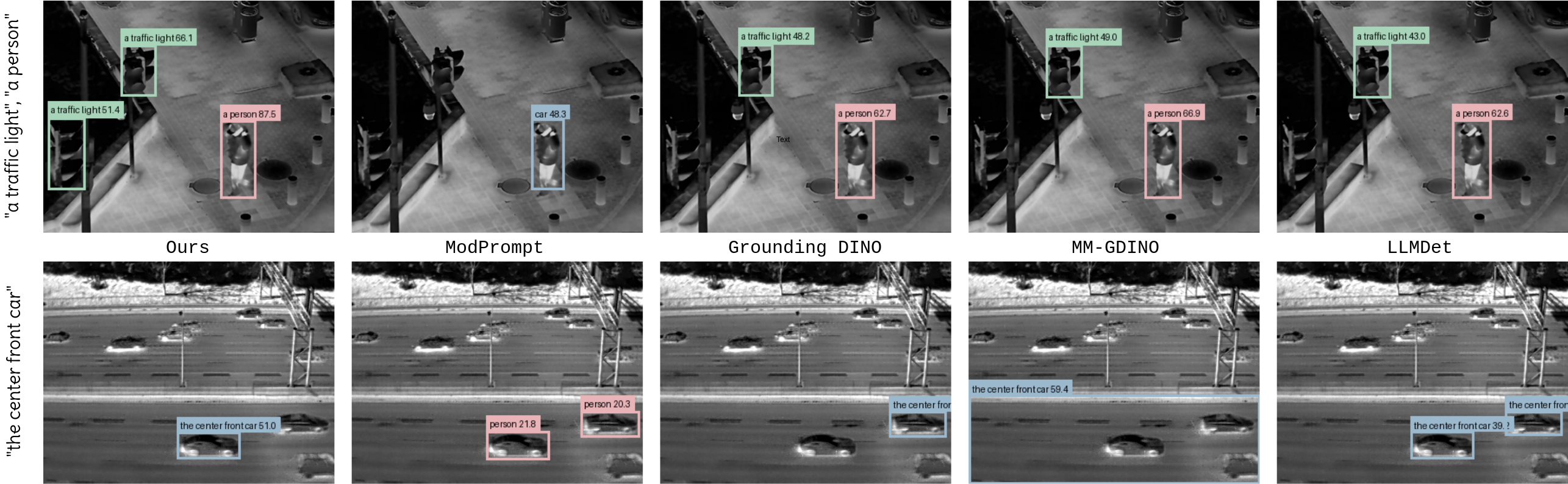}
    % \caption{Enter Caption}
    \captionof{figure}{Our method preserves full open-vocabulary reasoning in the thermal domain, outperforming both RGB-trained open-vocabulary detectors and adaptation-based thermal models.
In the top row, our detector produces higher-confidence predictions and correctly handles edge-case text queries that RGB models misdetect and adaptation approaches cannot interpret due to their loss of text-embedding generality.
In the bottom row, our model uniquely supports location-referenced detection (e.g., “the center front car”), a capability unreliably transferred from RGB detectors and entirely absent in adaptation-based pipelines.
These results highlight our framework’s ability to maintain semantic flexibility, robust grounding, and referential understanding under thermal domain shift.
    }
    \label{fig:teaser}
  \end{center}
  \vspace{0.5em} % small space before abstract; tweak as needed
}]

\begin{abstract}
Existing open-vocabulary detectors focus on RGB images and fail to generalize to thermal imagery, where low texture and emissivity variations challenge RGB-based semantics. We present \textit{\method}, the first large language model (LLM) supervised open-vocabulary detector tailored for thermal images. To enable large-scale training, we develop a synthetic dataset by converting GroundingCap-1M into the thermal domain and filtering captions to remove RGB-specific terms, yielding over one million thermally aligned samples with bounding boxes, grounding texts, and detailed captions. 
\method\ jointly optimizes detection, captioning, and cross-modal distillation objectives. A frozen RGB teacher provides geometric and semantic pseudo-supervision for paired but unlabeled RGB–thermal data, transferring open-vocabulary knowledge without manual annotation. The model further employs a \textit{Thermal–Text Alignment Head} for text calibration and a \textit{Modality-Fused Cross-Attention} module for dual-modality reasoning. Unlike prior domain-adaptation methods, the detector is fully fine-tuned to internalize thermal contrast patterns while preserving language alignment. 
Experiments on public benchmarks show consistent \textit{2–4\% AP gains} over existing open-vocabulary detectors, establishing a strong foundation for scalable, language-driven thermal perception.
\end{abstract}

\begin{figure*}[!ht]
    \centering
    \includegraphics[width=.95\linewidth]{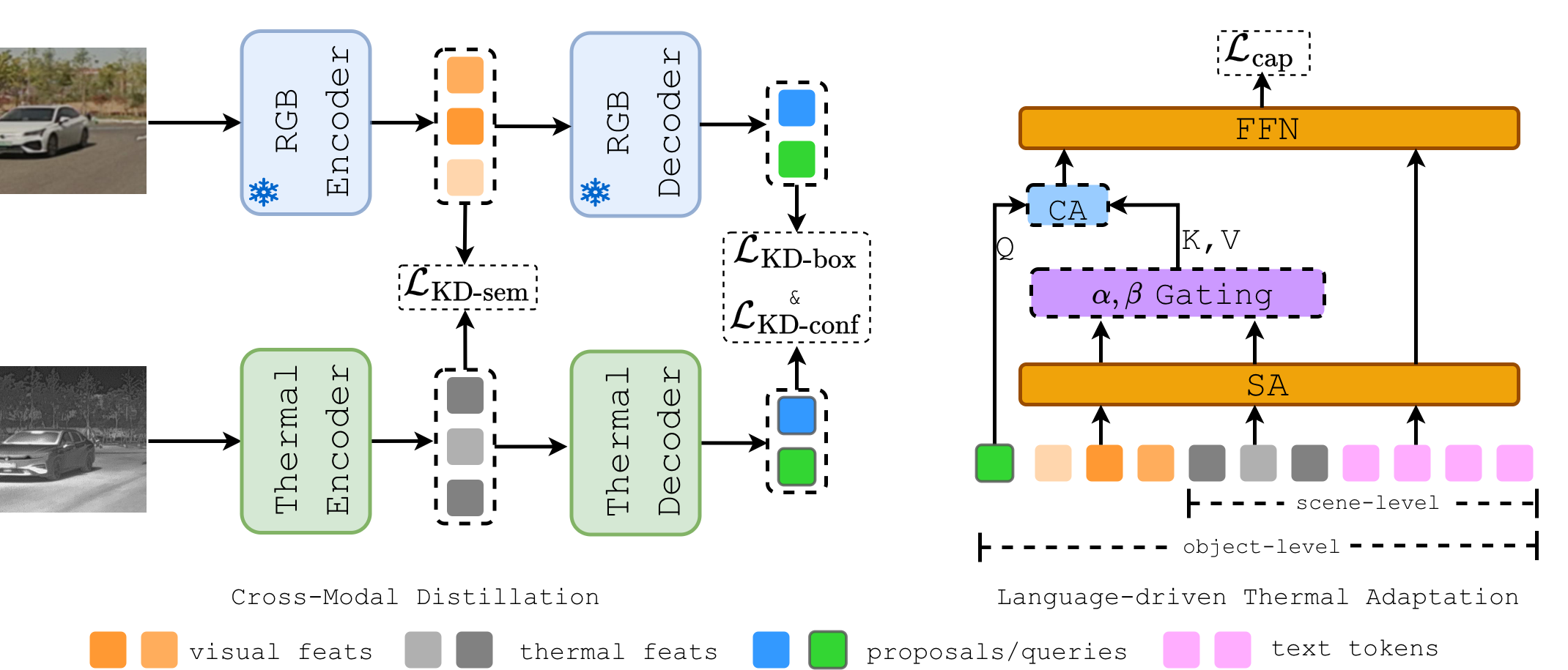}
\vskip-2pt\caption{\method\ combines a large language model (LLM) with a dual-stream RGB--thermal detector for zero-shot thermal object detection. 
The RGB branch acts as a frozen teacher, while the thermal branch learns transferable semantics through feature-, box-, and confidence-level distillation losses (\(\mathcal{L}_{\text{KD-sem}}, \mathcal{L}_{\text{KD-box}}, \mathcal{L}_{\text{KD-conf}}\)). 
The LLM generates scene- and object-level captions from visual proposals and text tokens, guided by a caption loss (\(\mathcal{L}_{\text{cap}}\)) with gating-controlled cross-attention. 
The LLM is discarded at inference.
% , adding no extra computational cost.
}
    \label{fig:pipeline}
\end{figure*}

\section{Introduction}

Thermal imaging is vital in safety-critical domains such as autonomous driving \cite{gade2014thermal, jiang2022object, leira2021object}, surveillance \cite{yeom2024thermal, jones2001novel}, and search-and-rescue \cite{castro2016thermal, rodin2018object}.
% , because it captures emitted infrared radiation rather than reflected light.
% , enabling reliable perception in darkness, fog, and smoke. 
% This makes thermal cameras effective for detecting humans and heat-emitting objects and a strong complement to RGB sensors under adverse conditions. 
However, progress in high-level thermal understanding, particularly object detection, remains limited by scarce annotated datasets and small benchmarks, leaving existing models restricted to a few predefined categories and unable to generalize to open-world cases.

Existing thermal object detectors rely heavily on supervised learning using datasets such as CAMEL \cite{gebhardt2018camel}, FLIR \cite{FLIR_Thermal_Dataset_2019}, LLVIP \cite{jia2021llvip, https://doi.org/10.48550/arxiv.2108.10831}, and KAIST \cite{choi2018kaist}, where annotations are limited to a few object types such as pedestrians and vehicles. While these methods \cite{medeiros2025visual, bahng2022exploring, jia2022visual, zhou2022conditional, zhou2022learning} achieve strong performance within their training domains, they lack the scalability to recognize novel or fine-grained categories due to the high cost and difficulty of labeling thermal imagery as illustrated in \cref{fig:teaser}. Recent efforts have explored domain adaptation \cite{medeiros2025visual, vs2022meta, guo2019domain, munir2021sstn} and synthetic data generation \cite{FViTA, xiao2025thermalgen, horstmann2025inference} to mitigate this limitation. However, these approaches primarily operate at the pixel or feature level, neglecting the higher-level semantic alignment necessary for open-world understanding. As a result, current thermal detectors remain closed-set, struggling to generalize to unseen object classes or adapt to real-world scenarios where new object types frequently appear.

Open-vocabulary object detection (OVD) has recently emerged as a powerful paradigm for scalable recognition, driven by models such as GLIP \cite{li2022grounded}, Grounding DINO \cite{liu2024grounding}, OWLv2 \cite{minderer2023scaling}, and LLMDet \cite{fu2025llmdet} that couple visual and linguistic supervision. These models leverage large-scale RGB datasets with paired captions to learn rich, text-conditioned visual representations, enabling detection of unseen categories through natural-language prompts. However, their success is largely confined to the visible spectrum, where semantics and texture cues are consistent with pretraining data \cite{yao2021filip, pan2022contrastive}. When transferred directly to thermal imagery, their performance deteriorates sharply due to the modality gap as depicted in \cref{fig:teaser}. Recent thermal extensions of open-vocabulary models attempt to bridge this gap by adapting RGB-pretrained detectors via parameter-efficient fine-tuning \cite{hu2022lora, fu2023effectiveness, liu2025pm, rauch2021object} or adapter-based training \cite{yin2023adapter, liu2025pm} on thermal datasets. While these approaches improve domain alignment, they still depend on limited thermal annotations and fail to generalize to unseen objects.
% , leaving open-vocabulary detection in the thermal domain an unsolved challenge.

To address these limitations, we introduce the first zero-shot open-vocabulary detection framework for thermal imagery, designed to operate without any thermal annotations. Our approach combines three complementary components that together enable scalable, label-free thermal perception. First, we construct a large-scale synthetic thermal dataset by converting GroundingCap-1M into the thermal domain using the F-ViTA translation model, providing diverse grounding and caption pairs under realistic thermal conditions. Second, we propose the Thermal-Text Alignment Head (TTAH), a lightweight module that recalibrates CLIP text embeddings toward thermal semantics by integrating subclass-level radiometric cues. Finally, we utilize RGB to thermal knowledge distillation, which transfers detection ability from a frozen RGB teacher to a thermal student using paired RGB-thermal datasets, allowing the model to inherit open-vocabulary reasoning from visible-spectrum detectors while adapting to thermal contrast patterns. Together, these components enable thermal detectors to recognize unseen object categories through text prompts without any manually annotated thermal data.

Experiments on public benchmarks show that our model consistently outperforms current open-vocabulary detectors, achieving between $2-4\%$ AP gains in zero-shot settings. The results demonstrate robust generalization and strong alignment between thermal and textual representations without any thermal labels. In summary, \method\ unifies synthetic supervision, semantic alignment, and cross-modal distillation to enable the first annotation-free, language-driven thermal detector for perception.

\section{Synthetic Thermal Dataset Construction}
\label{sec: synth dataset}

To enable large-scale pretraining of the thermal open-vocabulary detector, we construct a synthetic thermal dataset by converting the RGB images of the GroundingCap-1M dataset into their thermal-domain counterparts. 
GroundingCap-1M provides more than one million training samples of the form $(I_{\text{rgb}}, T_g, B, T_c)$, where $I_{\text{rgb}}$ is the RGB image, $T_g$ is the grounding text associated with object-level bounding boxes $B$, and $T_c$ is a detailed scene-level caption. 
The dataset integrates annotations from multiple large-scale sources such as COCO \cite{lin2014microsoft}, V3Det \cite{wang2023v3det}, GoldG \cite{kebe2021a}, and LCS-558k \cite{liu2023visual}, and its captions were originally generated using the Qwen2-VL-72B vision–language model to ensure factual, information-dense descriptions. 
We leverage this richly annotated corpus as the linguistic and geometric foundation for synthetic thermal supervision.

Synthesizing thermal data from GroundingCap-1M provides two key benefits. 
First, it enables large-scale pretraining of the thermal detector under rich linguistic supervision, covering over 13k categories inherited from V3Det and other open-vocabulary datasets, far exceeding the label diversity of existing thermal datasets such as KAIST, FLIR-ADAS, or LLVIP. 
Second, it offers a geometrically consistent framework for initializing the detector before cross-modal distillation, allowing the model to internalize thermal appearance priors without requiring expensive manual annotation. 
This synthetic dataset thus serves as the foundation for subsequent adaptation and knowledge transfer to real thermal imagery.\\
%\paragraph{Thermal Image Synthesis.}
\noindent{\bf{Thermal Image Synthesis.}}
To translate RGB imagery into the thermal domain while preserving scene structure and object geometry, we employ F-ViTA, a cross-domain image translation model trained for RGB-to-infrared conversion. 
Each RGB image $I_{\text{rgb}}$ in GroundingCap-1M is passed through F-ViTA to produce a synthetic thermal image $I_{\text{th}}^{\text{syn}}$. 
This process generates pixel-level thermal analogs that maintain the spatial alignment of original bounding boxes $B$, allowing direct reuse of existing annotations. 
The resulting dataset thus retains the original supervision structure $(I_{\text{th}}^{\text{syn}}, T_g, B, T_c)$ but operates fully in the thermal domain.\\ 
% Images are saved in 8-bit grayscale or pseudo-colored thermal format to ensure compatibility with downstream detection pipelines.
%\paragraph{Caption Adaptation.}
\noindent{\bf{Caption Adaptation.}}
Since the original captions $T_c$ in GroundingCap-1M were written for natural RGB scenes, we apply a lightweight text filtering step to remove RGB-specific descriptors such as color terms (\textit{red, blue, green}) or lighting conditions (\textit{bright, shadowed, sunlit}). 
This ensures that captions remain semantically faithful and visually grounded in the thermal representation while preserving object categories, spatial relations, and contextual cues. 
The grounding texts $T_g$ and bounding boxes $B$ are reused without modification, providing dense supervision for both object-level and scene-level objectives.

% \clearpage
\section{Language-Guided Training for Thermal Open-Vocabulary Detection}
\label{sec: method}

% Thermal object detection presents fundamental challenges for open-vocabulary models that are primarily trained on RGB imagery. Differences in texture, emissivity, and spectral response cause significant domain shifts, while the scarcity of annotated thermal data limits the ability of large detectors to adapt. To overcome these limitations, we introduce a unified framework for \textbf{thermal open-vocabulary detection} that combines full fine-tuning of a vision-language detector with cross-modal knowledge distillation and language-guided thermal adaptation. Our approach integrates three complementary learning signals:

% Thermal object detection presents fundamental challenges for open-vocabulary models that are primarily trained on RGB imagery. Differences in texture, emissivity, and spectral response cause significant domain shifts, while the scarcity of annotated thermal data limits the ability of large detectors to adapt. se limitations, we introduce a unified framework for \textbf{thermal open-vocabulary detection} 

To overcome the limitations thermal object detection presents for open-vocabulary models, we introduce a unified framework that combines full fine-tuning of a vision-language detector with cross-modal knowledge distillation and language-guided thermal adaptation. Our approach integrates complementary learning signals:

\begin{enumerate}
    % \item \textbf{Synthetic Thermal Supervision.} We generate a large-scale synthetic thermal dataset by translating RGB grounding images into the thermal domain while preserving their bounding boxes and captions. This dataset provides dense, supervised grounding signals for initializing the detector in a thermally consistent visual space.
    
    \item \textbf{Cross-Modal Distillation.} Using paired RGB-thermal datasets, a frozen RGB teacher detector transfers both spatial and semantic knowledge to the thermal student through box-, embedding-, and confidence-level distillation. This step bridges the appearance gap between synthetic and real thermal scenes without requiring any thermal annotations.
    
    \item \textbf{Language-Driven Thermal Adaptation.} A large language model (LLM) is integrated with the detector to generate both object- and scene-level captions. The LLM is augmented with \textit{thermal adapters}, lightweight and trainable modules inserted within its transformer blocks, to encode temperature-specific semantics and align linguistic representations with thermal visual cues. 
\end{enumerate}

Together, these components form an end-to-end system capable of performing zero-shot detection and descriptive understanding on real thermal imagery without using thermal labels. An overview of the framework is illustrated in \cref{fig:pipeline}, highlighting the interactions among the thermal detector, RGB teacher, and thermal-adaptive LLM.

% \begin{figure}
%     \centering
%     \includegraphics[width=0.99\linewidth]{aainr.png}
%     \caption{Enter Caption}
%     \label{fig:overview}
% \end{figure}

% \section{Thermal-Aware Open-Vocabulary Detector}

\subsection{Base Detection Framework}

Our detector follows an open-vocabulary formulation \cite{liu2024grounding, cheng2024yolo, zhang2022dino} that jointly addresses object localization and language grounding \cite{thomason2022language, steels2012language} in the thermal domain. Given a thermal image $I_{\text{th}}$ and textual queries $\{t_c\}$ encoded by the frozen CLIP text encoder, the model predicts bounding boxes $\{B_i\}$ and similarity scores $\{s_i\}$ through a transformer-based query–key–value decoding process. 
Unlike adapter-based transfer approaches, we fully fine-tune the detector so that both convolutional and attention layers adapt to thermal-specific cues while remaining aligned with the fixed CLIP text space. The model is optimized using a standard detection objective:
\begin{equation}
\mathcal{L}_{\text{det}} = \mathcal{L}_{\text{cls}} + \mathcal{L}_{\text{box}}.
\end{equation}
Here, $\mathcal{L}_{\text{cls}}$ enforces vision–language alignment via cosine-similarity contrastive learning between region features and text embeddings, while $\mathcal{L}_{\text{box}}$ combines $\ell_1$ and GIoU/CIoU losses for precise localization. Training on the synthetic thermal dataset provides a strong initialization linking thermal features with caption-aligned semantic supervision, enabling effective downstream distillation and language-guided adaptation.

\subsection{Supervision of Large Language Models}

While full fine-tuning enables the detector to learn thermal-specific features, visual supervision alone cannot provide the compositional reasoning or linguistic understanding needed for open-vocabulary detection. Thermal images often contain weak textures, low contrast, and temperature-driven ambiguity, making category discrimination difficult without additional semantic cues. To introduce richer language context, we pair the detector with an LLM that generates object- and scene-level captions conditioned on thermal features. This coupling allows the LLM to ground natural language in thermally distinctive structures, strengthening cross-modal alignment and enhancing the model’s ability to reason about novel or fine-grained categories.

% \subsection{Architecture}

\noindent{\bf{LLM thermal adapters}}: The LLM receives projected feature tokens from both the thermal detector and, when available, the RGB teacher. A lightweight projector maps detector embeddings into the LLM token space. Inside the LLM, we insert \textit{Thermal Adapters}, small LoRA-style \cite{liu2024dora, hu2022lora} residual MLPs, into the feed-forward sub-layers of each transformer block.
% :
% \begin{equation}
% H_{\text{out}} = H_{\text{in}} + \gamma\, W_2(\sigma(W_1 H_{\text{in}})),
% \end{equation}
% where $H_{\text{in}}$ and $H_{\text{out}}$ denote the input and output hidden states, $W_1$ and $W_2$ are adapter weights, $\sigma(\cdot)$ is a GELU activation, and $\gamma$ controls the adapter strength. 
These adapters specialize the LLM to thermal-specific semantics, enabling effective transfer of textual priors to the thermal domain without disturbing the original linguistic knowledge.

% \subsection{Dual-Modality Cross-Attention}
\noindent{\bf Dual-Modality Cross-Attention:}
To strengthen object-level reasoning, we incorporate a \textit{Modality-Fused Cross-Attention} (MFCA) module in the caption head. MFCA allows LLM text queries $Q$ to attend jointly to thermal and RGB teacher features, leveraging the structural clarity of thermal imagery and the semantic richness of RGB signals. Given thermal and RGB key–value pairs, MFCA forms:
\begin{equation}
K = [\alpha K_{\text{th}};\, \beta K_{\text{rgb}}], \qquad
V = [\alpha V_{\text{th}};\, \beta V_{\text{rgb}}],
\end{equation}
where $\alpha$ and $\beta$ are learnable gates controlling each modality’s contribution. This fused attention encourages object-level captions that are semantically grounded while remaining faithful to thermal appearance cues. At inference, RGB inputs are absent; MFCA automatically collapses to thermal-only attention by setting $\beta = 0$, preserving semantic reasoning learned during training while enabling consistent captioning and detection performance on thermal data.

% An optional attention-alignment regularizer further ensures balanced utilization of modalities:
% \begin{equation}
% \mathcal{L}_{\text{CA-align}} = \| \bar{A}_{\text{th}} - \bar{A}_{\text{rgb}} \|_1,
% \end{equation}
% where $\bar{A}_{\text{th}}$ and $\bar{A}_{\text{rgb}}$ denote the mean attention maps across heads for the two modalities.

\subsection{Caption Generation Loss}
The proposed framework supervises the LLM with two complementary captioning objectives: a \textit{scene-level caption generation} loss for global scene understanding and an \textit{object-level caption generation} loss for fine-grained grounding.
% Both objectives follow the standard next-token prediction paradigm but differ in the nature of their visual inputs and supervision signals.
\\
%\paragraph{Scene-Level Caption Generation.}
\noindent{\bf Scene-Level Caption Generation.}
For scene-level supervision, the LLM receives detector-extracted thermal feature maps and is trained to generate long, descriptive captions for synthetic thermal images derived from GroundingCap-1M. The input is formatted conversationally, with a system instruction, a prompt (e.g., “Describe the thermal image in detail”), and the ground-truth caption as the expected response. Because the RGB teacher is not used here, the LLM learns to rely entirely on thermal cues, enabling the LLM and TTAH to embed radiometric structure, emissivity patterns, and global scene context into the language space.\\
%\paragraph{Object-Level Caption Generation.}
\noindent{\bf Object-Level Caption Generation.}
While scene-level captions promote global understanding, they offer limited grounding for specific regions. To align textual tokens with localized entities, we introduce an object-level captioning objective. For the \textit{synthetic thermal} data, we use the short region phrases provided by GroundingCap-1M (e.g., “man walking,” “car headlights”). For \textit{real paired} datasets such as M\textsuperscript{3}FD, which lack captions, we derive pseudo-phrases either from the RGB teacher’s open-vocabulary predictions or from dataset class labels (e.g., “person,” “car”). These phrases supervise the LLM to associate detector queries with meaningful linguistic concepts even without human annotations.
% Object-query features interact with detector visual tokens through our MFCA, which fuses thermal and RGB teacher cues:
% \begin{equation}
% K = [\alpha K_{\text{th}};\, \beta K_{\text{rgb}}], \qquad
% V = [\alpha V_{\text{th}};\, \beta V_{\text{rgb}}],
% \end{equation}
% with learnable gates $\alpha$ and $\beta$ modulating each modality. This enables RGB-enhanced semantic grounding during training, while naturally reducing to thermal-only attention ($\beta = 0$) at inference.
% The language modeling loss for object-level captioning is:
% \begin{equation}
% \mathcal{L}{\text{cap-region}} = -\sum{t=1}^{T} \log P(y_t \mid y_{<t}, F_{\text{th}}, F_{\text{rgb}}),
% \end{equation}
% applied only to paired RGB–thermal batches.
%\paragraph{Combined Objective.}
% \noindent{\bf{Combined Objective.}}
The total captioning loss combines both supervision levels:
\begin{equation}
\mathcal{L}_{\text{cap}} = \mathcal{L}_{\text{cap-scene}} + \mathcal{L}_{\text{cap-object}}.
\end{equation}
These objectives jointly teach the LLM to blend global thermal understanding with localized semantic grounding. 
% The scene-level loss captures holistic thermal context, while the object-level loss anchors language to specific regions via dual-modality attention, enabling reliable grounding even when RGB inputs are unavailable at inference.

\subsection{Cross-Modal Distillation}

The preceding modules enhance the model’s linguistic grounding and thermal–text alignment through caption supervision and the TTAH.
However, thermal imagery still exhibits a substantial domain gap in visual appearance and feature statistics relative to RGB data.
To bridge this gap, we introduce a cross-modal distillation framework in which a frozen RGB detector serves as a teacher and the thermal detector acts as a student.
This strategy transfers spatial and semantic knowledge from large-scale RGB-trained models to the thermal domain, enabling the detector to acquire open-vocabulary recognition capabilities without requiring manual thermal annotations.\\
%\paragraph{Teacher-Student Setup.}
\noindent{\bf{Teacher-Student Setup.}}
Given paired RGB–thermal frames $(I_{\text{rgb}}, I_{\text{th}})$ that are spatially aligned at the frame or pixel level (e.g., from M\textsuperscript{3}FD), the RGB teacher processes $I_{\text{rgb}}$ to produce pseudo-labels, including predicted bounding boxes, category logits, and object–text similarity scores.
The thermal student processes the corresponding thermal image $I_{\text{th}}$ and predicts its own detections.
Knowledge transfer occurs by minimizing the discrepancy between teacher and student predictions across three complementary dimensions-spatial geometry, visual semantics, and confidence calibration.
% \paragraph{Overall Objective.}
The total distillation loss is expressed as:
\begin{equation}
\mathcal{L}_{\text{KD}} = \mathcal{L}_{\text{KD-box}} + \mathcal{L}_{\text{KD-sem}} + \mathcal{L}_{\text{KD-conf}}.
\end{equation}
Each term governs a distinct aspect of alignment between the modalities.\\
%\paragraph{Spatial Distillation.}
\noindent{\bf{Spatial Distillation.}}
To ensure geometric consistency, the student’s predicted bounding boxes $B_{\text{th}}$ are aligned with the teacher’s boxes $B_{\text{rgb}}$ using a GIoU-based objective:
\begin{equation}
\mathcal{L}_{\text{KD-box}} = 1 - \text{GIoU}(B_{\text{rgb}}, B_{\text{th}}).
\end{equation}
This encourages the thermal detector to reproduce the teacher’s spatial localization patterns, preserving object boundaries despite differences in contrast and texture.\\
%\paragraph{Semantic Distillation.}
\noindent{\bf{Semantic Distillation.}}
High-level visual semantics are transferred by aligning feature embeddings from the teacher and student.
Given teacher region features $f_{\text{rgb}}$ and student features $f_{\text{th}}$, a cosine-based InfoNCE loss promotes modality-invariant representations:
\begin{equation}
\mathcal{L}_{\text{KD-sem}} = -\log 
\frac{
\exp(\cos(f_{\text{th}}, f_{\text{rgb}})/\tau)
}{
\sum_{j} \exp(\cos(f_{\text{th}}, f^{j}_{\text{rgb}})/\tau)
},
\end{equation}
where $\tau$ is a temperature coefficient.
This alignment allows the thermal student to inherit semantic abstraction capabilities from the RGB teacher while remaining grounded in its own radiometric appearance cues.\\
%\paragraph{Confidence Distillation.}
\noindent{\bf{Confidence Distillation.}}
To capture the teacher’s uncertainty and class probability structure, we minimize the Kullback-Leibler divergence between the normalized class-probability distributions:
\begin{equation}
\mathcal{L}_{\text{KD-conf}} = \mathrm{KL}(p_{\text{rgb}} \parallel p_{\text{th}}),
\end{equation}
where $p_{\text{rgb}}$ and $p_{\text{th}}$ represent the teacher’s and student’s class logits, respectively.
This encourages the student to emulate the teacher’s soft decision boundaries, improving robustness under ambiguous or low-contrast thermal conditions.

% \paragraph{Training Behavior.}

The distillation losses are activated only for paired RGB–thermal batches.
Synthetic thermal batches or thermal-only captioning samples bypass this module during training.
By combining these objectives, the thermal detector learns to replicate both the spatial precision and the semantic richness of the RGB teacher, effectively narrowing the visual domain gap while maintaining fidelity to the thermal modality.

\begin{table*}[h!]
\centering
\caption{Zero-shot detection results on thermal datasets FLIR-Aligned, FLIR-V2, camel, and Utokyo.}
\setlength{\tabcolsep}{4.7pt} % tighter column spacing
\renewcommand{\arraystretch}{1.0} % slightly tighter row spacing
\begin{tabular}{@{}lcccc|ccc|ccc|ccc@{}}
\toprule
\multirow{2}{*}{Method} & \multirow{2}{*}{Venue} &
\multicolumn{3}{c|}{FLIR-Aligned} &
\multicolumn{3}{c|}{FLIR-V2} &
\multicolumn{3}{c|}{CAMEL} &
\multicolumn{3}{c}{Utokyo} \\
\cmidrule(lr){3-14}
 & & AP & AP$_{50}$ & AP$_{75}$ & AP & AP$_{50}$ & AP$_{75}$ & AP & AP$_{50}$ & AP$_{75}$ & AP & AP$_{50}$ & AP$_{75}$ \\
\midrule
GLIP & NeurIPS'22 & 0.251 & 0.471 & 0.226 & 0.025 & 0.041 & 0.028 & 0.186 & 0.324 & 0.242 & 0.049 & 0.093 & 0.046 \\
T-Rex2 & ECCV'24 & 0.276 & 0.514 & 0.255 & 0.033 & 0.048 & 0.036 & 0.213 & 0.345 & 0.274 & 0.049 & 0.101 & 0.043 \\
YOLO-World & CVPR'24 & 0.266 & 0.487 & 0.241 & 0.029 & 0.044 & 0.032 & 0.197 & 0.336 & 0.256 & 0.050 & 0.101 & 0.043 \\
G-DINO & ECCV'24 & 0.337 & \underline{0.636} & 0.313 & \underline{0.081} & \underline{0.144} & \underline{0.078} & \underline{0.482} & \underline{0.729} & \textbf{0.543} & \underline{0.050} & 0.093 & 0.169 \\
MM-GDINO & -- & 0.354	& 0.619 & \textbf{0.371} & 0.042 & 0.060 &	0.048 & 0.273	& 0.369 & 0.340 & 0.047 & 0.095 & 0.042 \\
LLMDet & CVPR'25 & \underline{0.359} & 0.628 & 0.348 & 0.048 & 0.075 & 0.051 & 0.383 & 0.560 & 0.439 & 0.050 & 0.102 & 0.045 \\
\textbf{Ours} & -- & \textbf{0.372} & \textbf{0.664} & \underline{0.359} & \textbf{0.096} & \textbf{0.173} & \textbf{0.091} & \textbf{0.511} & \textbf{0.758} & \underline{0.525} & \textbf{0.065} & \textbf{0.137} & \textbf{0.054} \\
\bottomrule
\end{tabular}
\label{tab:thermal metrics}
\end{table*}

\subsection{Thermal-Text Alignment Head (TTAH)}

Although the detector and LLM are fine-tuned on thermal data, the textual embeddings from the frozen CLIP encoder remain biased toward RGB visual statistics. This bias introduces a semantic gap between thermally adapted visual features and text representations that were learned predominantly from natural-light imagery. 

To address this gap, we propose the \textit{Thermal-Text Alignment Head} (TTAH), a lightweight module that recalibrates CLIP text embeddings toward the thermal feature space while preserving compatibility with the original CLIP embedding manifold. TTAH operates exclusively within the CLIP text branch and is active for all thermal-text similarity computations used in detection and caption supervision.

% \paragraph{Design.}
For each text token embedding $t_c \in \mathbb{R}^d$ produced by the frozen CLIP encoder, TTAH augments it with a small radiometric attribute vector $a_j$ drawn from a learnable bank of thermal descriptors (e.g., \textit{hot}, \textit{silhouette}, \textit{reflective}, \textit{high-emissivity}). The concatenated representation is transformed by a two-layer MLP followed by layer normalization:
\begin{equation}
t_c^{*} = \mathrm{LN}\!\left(\mathrm{MLP}\!\left([\; t_c ; a_j \;]\right)\right),
\end{equation}
where $[\,\cdot\,;\,\cdot\,]$ denotes vector concatenation. The resulting thermally calibrated embeddings $t_c^{*}$ replace $t_c$ in all downstream similarity computations, including object-text contrastive losses and caption alignment objectives.

% \paragraph{Learning Objective.}
TTAH is trained jointly with the rest of the model using batches that include paired thermal-text supervision. It is optimized with an object-text contrastive loss that aligns thermal visual embeddings $f_{\text{th}}$ with their thermally adjusted text embeddings $t_c^{*}$:
\begin{equation}
\mathcal{L}_{\text{TTAH-ctr}} = -\log 
\frac{
\exp(\cos(f_{\text{th}}, t_c^{*}) / \tau)
}{
\sum_{k} \exp(\cos(f_{\text{th}}, t_k^{*}) / \tau)
},
\end{equation}
where $\tau$ is a temperature coefficient controlling contrastive sharpness. To prevent semantic drift from the original CLIP space, a regularization term constrains the calibrated text embeddings to remain close to their original representations:
\begin{equation}
\mathcal{L}_{\text{TTAH-drift}} = \|\, t_c^{*} - t_c \,\|_2^2.
\end{equation}
The overall TTAH loss combines both terms:
\begin{equation}
\mathcal{L}_{\text{TTAH}} = \mathcal{L}_{\text{TTAH-ctr}} + \lambda_{\text{drift}}\,\mathcal{L}_{\text{TTAH-drift}},
\end{equation}
where $\lambda_{\text{drift}}$ balances alignment strength and stability.

% \paragraph{Attribute-Expanded Labeling.}
To adapt text embeddings dynamically to varying thermal conditions, TTAH extends each base class label into multiple \textit{thermal sublabels} by pairing it with every radiometric attribute vector in the learned attribute bank $\{a_1, a_2, \ldots, a_M\}$. For a class $c$ with frozen CLIP embedding $t_c$, each sublabel embedding is generated as:
\begin{equation}
t_{c,j}^{*} = \mathrm{LN}\!\left(\mathrm{MLP}\!\left([\; t_c ; a_j \;]\right)\right), \quad j = 1, \ldots, M.
\end{equation}
Given a thermal region feature $f_{\text{th}}$, similarity scores are computed between the visual embedding and all $M$ thermal variants of class $c$:
\begin{equation}
s_{c,j} = \cos(f_{\text{th}}, t_{c,j}^{*}),
\end{equation}
and the most compatible variant is selected as the effective text representation:
\begin{equation}
\tilde{t}_c = t_{c, j^{*}(c)}^{*}, \quad j^{*}(c) = \arg\max_{j} s_{c,j}.
\end{equation}
The final classification score for class $c$ is then $\hat{s}_c = \max_j s_{c,j}$, and the predicted label is $\hat{y} = \arg\max_c \hat{s}_c$. 

This attribute expansion allows each base label to adapt semantically to the thermal characteristics of a scene, e.g., ``hot person,'' ``silhouette vehicle,'' or ``reflective surface'', without requiring any additional annotations. During training, supervision remains at the base-label level, enabling the model to implicitly infer which thermal variant best aligns with each instance. The same mechanism applies at inference, where all sublabels are evaluated per class to determine the most thermally consistent representation.

% \paragraph{Behavior and Role.}
TTAH operates solely on text embeddings associated with thermal imagery; RGB features and RGB-generated captions are excluded from its optimization. This targeted alignment ensures that textual representations adapt specifically to the thermal feature distribution rather than inheriting RGB biases. 

By introducing radiometric attributes as auxiliary conditioning signals, TTAH learns to link linguistic cues such as \textit{heat source}, \textit{cold region}, and \textit{silhouette} with distinctive thermal structures. These calibrated text embeddings improve both open-vocabulary detection and caption generation by providing a language space that is semantically aligned and radiometrically consistent with thermal vision.

\subsection{Objective Formulation}

The proposed framework integrates multiple learning signals originating from detection supervision, cross-modal distillation, thermally adaptive text alignment, and language-based caption generation. All objectives are optimized jointly within a single end-to-end training pipeline. The overall objective is expressed as:
\begin{equation}
\mathcal{L}_{\text{total}} = 
\mathcal{L}_{\text{det}} +
\mathcal{L}_{\text{KD}} +
% \mathcal{L}_{\text{KD-sem}} +
% \mathcal{L}_{\text{KD-conf}} +
\mathcal{L}_{\text{TTAH}} +
\mathcal{L}_{\text{cap}}.
\end{equation}

Each component plays a distinct role in ensuring effective transfer, grounding, and generalization.
$\mathcal{L}_{\text{det}}$ supervises the detector using fully labeled synthetic thermal data; $\mathcal{L}_{\text{KD}}$ enforces geometric, semantic, and probabilistic alignment between the RGB teacher and the thermal student; $\mathcal{L}_{\text{TTAH}}$ bridges the domain gap between CLIP text embeddings and thermal representations; and $\mathcal{L}_{\text{cap}}$ trains the thermal-adaptive LLM to generate coherent scene- and object-level captions aligned with visual cues.

The combination of these terms ensures that the model learns to perceive, localize, and linguistically describe thermal scenes without direct supervision from thermal annotations.

% \clearpage
% \input{sec/results}
\section{Experiment}
\label{sec: experiment}

% Implementation details are provided in the supplementary material.

\subsection{Zero-Shot Detection Transfer Ability}

% We further assess the cross-domain generalization ability of our thermal open-vocabulary detector on three real infrared benchmarks: FLIR-Aligned, FLIR-V2, and CAMEL. 
% Importantly, no thermal annotations from these datasets are used during training; the model is trained only on synthetic thermal data and unlabeled paired RGB-thermal data via distillation.

We evaluate cross-domain generalization on seven real infrared benchmarks: FLIR-Aligned, FLIR-V2, CAMEL, SMOD, Utokyo, MFAD, and LLVIP. None provide thermal annotations during training; the model is trained only on synthetic thermal data from GroundingCap-1M and unlabeled paired RGB–thermal datasets via distillation. This setup tests true zero-shot transfer across diverse resolutions, sensors, and scene conditions.

\begin{figure}[!t]
  \centering
  % \fbox{\rule{0pt}{2in} \rule{3in}{0pt}}
  \includegraphics[width=.95\linewidth]{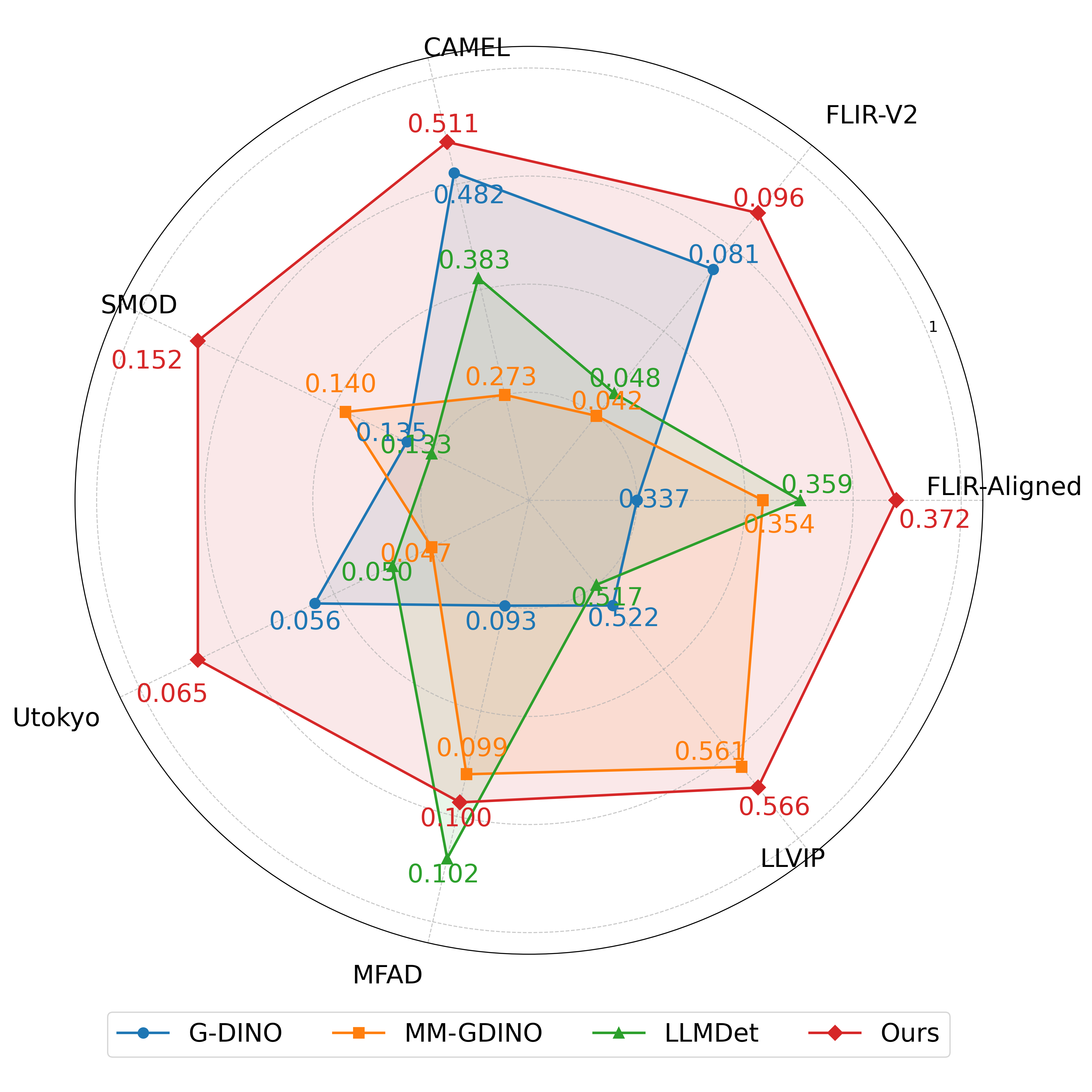}
 \vskip-12pt \caption{\method\ zero-shot performance across
various benchmarks compared with other OVDs with Swin-T backbone.}
  \label{fig: radar plot}
\end{figure}

As summarized in \cref{tab:thermal metrics} and visualized in \cref{fig: radar plot}, our method consistently outperforms RGB-based open-vocabulary detectors including GLIP, T-Rex2, YOLO-World, Grounding DINO (G-DINO), MM-GDINO, and LLMDet, all evaluated under the same Swin-T backbone. On FLIR-Aligned, the model achieves an AP of 0.372, improving over LLMDet (0.359) and MM-GDINO (0.354), and reaching the highest AP$_{50}$ of 0.664. On the more challenging FLIR-V2 benchmark, it attains an AP of 0.096, surpassing G-DINO (0.081) and LLMDet (0.048), and setting new highs for AP$_{50}$ (0.173) and AP$_{75}$ (0.091). On CAMEL, which contains dense and diverse multi-object scenes, the model obtains the top AP of 0.511, outperforming G-DINO (0.482) and LLMDet (0.383), along with the highest AP$_{50}$.

Further results across SMOD, Utokyo, MFAD, and LLVIP reinforce the model’s strong cross-domain generalization. On SMOD, it reaches an AP of 0.152, exceeding all RGB-based detectors, while on Utokyo it achieves 0.065 AP, improving over G-DINO (0.056). On MFAD, the model performs competitively at 0.100 AP, close to LLMDet (0.102) despite the absence of thermal annotations during training. On LLVIP, which involves challenging nocturnal scenes, the method delivers the best results with 0.566 AP and 0.856 AP$_{50}$. Together, these findings demonstrate that the combination of synthetic thermal supervision, thermal-text alignment, and RGB-to-thermal distillation enables effective transfer across diverse infrared domains under a true zero-shot setting.

These results show that RGB open-vocabulary detectors, even when trained on large grounding or captioning datasets, fail to transfer reliably to thermal imagery. In contrast, our method combines synthetic thermal supervision with RGB to thermal distillation, enabling the model to absorb both thermal-specific cues and rich semantic structure. This produces a detector that is modality aware, language aligned, and effective for zero-shot thermal detection.

\subsection{Class- and Scale-Wise Analysis on FLIR-V2}

We perform a class- and scale-wise breakdown on FLIR-V2 to characterize where our zero-shot transfer succeeds and where it falls short. For dominant thermal classes, the ZS transfer is strong: \emph{person} achieves 0.366 AP (AP$_{50}$=0.56, AP$_{75}$=0.40) versus 0.512 AP for the fully supervised G-DINO (FT) baseline, with competitive medium-scale performance (AP$_m$: 0.783 vs.\ 0.836), while \emph{car} reaches 0.35 AP with high large-object accuracy (AP$_l$=0.882 vs.\ 0.958 for FT). In contrast, errors concentrate on rare, small, and thermally ambiguous classes such as \emph{motorcycle} (0.006 AP) and \emph{traffic light} (0.006 AP), reflecting small-object localization limits rather than semantic collapse. These failure modes are dominated by small-instance localization—e.g., \emph{motorcycle} exhibits AP$_s$=0.001—and weak thermal contrast for fine-grained objects. Notably, similar trends persist even under full supervision, indicating that the remaining gap reflects inherent sensor and scale limitations rather than deficiencies in the zero-shot transfer mechanism itself.

\subsection{Evaluation Across Teacher Backbones}

We evaluate our framework using three RGB open-vocabulary detectors; Grounding DINO, MM-GDINO, and LLMDet, as teacher backbones. As shown in \cref{tab:flir_camel_results}, the proposed thermal adaptation and RGB to thermal distillation consistently improve detection accuracy across all architectures on both FLIR-Aligned and CAMEL datasets. MM-GDINO achieves the highest relative gain, with $+37.6\%$ AP on FLIR and $+15.2\%$ AP on CAMEL, suggesting that its broader RGB pretraining corpus (O365, GRIT, V3Det) provides a strong supervision signal for cross-modal transfer. Nevertheless, the adapted Grounding DINO backbone attains the highest absolute validation performance after thermal training and is therefore adopted for all subsequent evaluations. Larger improvements are observed on FLIR-Aligned, a purely thermal dataset, confirming that the proposed distillation and alignment strategy is particularly effective when texture cues are minimal. The consistent AP$_{75}$ gains further indicate that the Thermal-Text Alignment Head (TTAH) preserves fine-grained localization. Overall, these results demonstrate that our framework generalizes reliably across backbones and serves as a backbone-agnostic solution for thermal OVD.

\subsection{Effect of Each Component in the Pipeline}

We perform an ablation study on the FLIR-Aligned and CAMEL validation sets to evaluate the individual contributions of each component in our training framework. As shown in \cref{tab:ablation_components}, performance improves progressively as different objectives are added, demonstrating the complementary nature of the proposed components.

Starting from the zero-shot baseline, the model achieves 0.200 AP on FLIR-Aligned and 0.547 AP on CAMEL. Introducing the scene-level caption loss ($\mathcal{L}_{\text{cap-scene}}$) provides a modest gain of $+0.012$ AP on FLIR and $+0.006$ AP on CAMEL, helping the model capture global semantic consistency between captions and thermal imagery. 
When adding the object-level caption regularization ($\mathcal{L}_{\text{cap-object}}$), the performance increases substantially by $+0.028$ AP on FLIR and $+0.020$ AP on CAMEL. This demonstrates that local alignment between thermal object regions and caption phrases is more effective than global supervision, especially under low-texture and high-emissivity conditions.

\begin{table}[h!]
\centering
\caption{Comparison of detection performance on FLIR-Aligned and CAMEL datasets. Bold rows indicate our model’s performance relative to each baseline.}
\label{tab:flir_camel_results}
\small
\setlength{\tabcolsep}{5.7pt} % tighter column spacing
\begin{tabular}{@{}lccc|ccc@{}}
\toprule
\multirow{2}{*}{Method} & \multicolumn{3}{c|}{FLIR-Aligned} & \multicolumn{3}{c}{CAMEL} \\
\cmidrule(lr){2-7}
 & AP & AP$_{50}$ & AP$_{75}$ & AP & AP$_{50}$ & AP$_{75}$ \\
\midrule
GDINO & 0.200 & 0.435 & 0.164 & 0.547 & 0.781 & 0.657 \\
{Ours} & 0.261 & 0.571 & 0.207 & 0.585 & 0.802 & 0.707 \\
$\Delta~\%$ & 30.5 & 31.2 & 26.2 & 6.9 & 2.7 & 7.6 \\
\midrule
MM-GDINO & 0.170 & 0.348 & 0.145 & 0.492 & 0.681 & 0.599 \\
{Ours} & 0.234 & 0.510 & 0.188 & 0.567 & 0.788 & 0.677 \\
$\Delta~\%$ & 37.6 & 46.5 & 29.6 & 15.2 & 15.7 & 13.0 \\
\midrule
LLMDet & 0.201 & 0.394 & 0.171 & 0.552 & 0.765 & 0.670 \\
{Ours} & 0.255 & 0.548 & 0.207 & 0.574 & 0.817 & 0.694 \\
$\Delta~\%$ & 26.8 & 39.1 & 21.1 & 3.9 & 6.8 & 3.6 \\
\bottomrule
\end{tabular}
\end{table}

\begin{table}[h!]
\centering
\caption{Ablation study showing incremental contributions of each component on FLIR-Aligned and CAMEL datasets. Larger gains are observed from $\mathcal{L}_{\text{cap-reg}}$ and $\mathcal{L}_{\text{KD}}$.}
\label{tab:ablation_components}
\small
\begin{tabular}{@{}lccc|ccc@{}}
\toprule
\multirow{2}{*}{Component} & \multicolumn{3}{c|}{FLIR-Aligned} & \multicolumn{3}{c}{CAMEL} \\
\cmidrule(lr){2-7}
 & AP & AP$_{50}$ & AP$_{75}$ & AP & AP$_{50}$ & AP$_{75}$ \\
\midrule
Zero-shot & 0.200 & 0.435 & 0.164 & 0.547 & 0.781 & 0.657 \\
$+~\mathcal{L}_{\text{cap-img}}$ & 0.012 & 0.025 & 0.009 & 0.006 & 0.004 & 0.007 \\
$+~\mathcal{L}_{\text{cap-reg}}$ & 0.028 & 0.073 & 0.020 & 0.020 & 0.012 & 0.026 \\
$+~\mathcal{L}_{\text{KD}}$ & 0.021 & 0.038 & 0.014 & 0.012 & 0.005 & 0.017 \\
\midrule
$\Delta$ & 0.061 & 0.136 & 0.043 & 0.038 & 0.021 & 0.050 \\
Final & 0.261 & 0.571 & 0.207 & 0.585 & 0.802 & 0.707 \\
\bottomrule
\end{tabular}
\end{table}

\begin{table}[h!]
\centering
\caption{Ablation on subclass selection strategies within the TTAH module, evaluated on FLIR-Aligned and CAMEL datasets.}
\label{tab:ttah_subclass_selection}
\small
\begin{tabular}{@{}lccc|ccc@{}}
\toprule
\multirow{2}{*}{Strategy} & \multicolumn{3}{c|}{FLIR-Aligned} & \multicolumn{3}{c}{CAMEL} \\
\cmidrule(lr){2-7}
 & AP & AP$_{50}$ & AP$_{75}$ & AP & AP$_{50}$ & AP$_{75}$ \\
\midrule
Average  & 0.234 & 0.489 & 0.182 & 0.559 & 0.782 & 0.678 \\
Random & 0.248 & 0.528 & 0.196 & 0.573 & 0.794 & 0.690 \\
Conf. gating & 0.261 & 0.571 & 0.207 & 0.585 & 0.802 & 0.707 \\
\bottomrule
\end{tabular}
\end{table}

The knowledge distillation loss ($\mathcal{L}_{\text{KD}}$) further improves the model by $+0.021$ AP on FLIR and $+0.012$ AP on CAMEL, highlighting the critical role of paired RGB-thermal data. This component enables direct transfer of geometric and semantic knowledge from the RGB teacher, offering tangible improvements over relying solely on synthetic data. The combined use of caption regularization and distillation leads to cumulative gains of $+6.1\%$ AP on FLIR-Aligned and $+3.8\%$ AP on CAMEL compared to the zero-shot baseline.
The full model reaches 0.261 AP / 0.571 AP$_{50}$ / 0.207 AP$_{75}$ on FLIR-Aligned and 0.585 AP / 0.802 AP$_{50}$ / 0.707 AP$_{75}$ on CAMEL, confirming improvements in both recall (AP$_{50}$) and localization precision (AP$_{75}$). These results validate that each proposed component plays a distinct and complementary role: global captioning enhances contextual understanding, object-level supervision refines spatial grounding, and distillation leverages real paired data for robust thermal detection. To further isolate the contribution of TTAH, we remove $\mathcal{L}_{\text{TTAH}}$ while keeping all other components fixed; this reduces performance by 0.008 AP / 0.019 AP$_{50}$ / 0.006 AP$_{75}$ on FLIR-Aligned, confirming that TTAH provides complementary gains beyond captioning and distillation.

\subsection{Effect of Subclass Selection Strategies in TTAH}

We analyze the impact of different subclass selection strategies within the Thermal-Text Alignment Head (TTAH) on thermal-text alignment and downstream detection accuracy. As shown in \cref{tab:ttah_subclass_selection}, the strategy used to select subclass text embeddings substantially influences model performance. 
The \textit{average-pooling} baseline, which treats all subclass embeddings equally, performs the worst with 0.234 AP on FLIR-Aligned and 0.559 AP on CAMEL, indicating that uniform weighting weakens the contribution of discriminative subclass cues such as ``warm person'' or ``cold vehicle.'' 
\textit{Random sampling} of subclass prompts improves performance moderately, reaching 0.248 AP on FLIR and 0.573 AP on CAMEL, suggesting that stochastic subclass variation introduces some diversity but lacks stability. 
The proposed \textit{confidence-gated} strategy achieves the best results, 0.261 AP on FLIR and 0.585 AP on CAMEL, by weighting subclass embeddings according to their predicted text-image similarity. This adaptive mechanism consistently emphasizes semantically relevant subclasses and suppresses noisy or less informative ones, yielding $+2.7$ AP improvement over average pooling and $+1.3$ AP over random selection. The gain is more pronounced on FLIR-Aligned, where texture cues are limited, confirming that confidence-guided subclass selection strengthens cross-modal alignment and improves zero-shot detection in thermal imagery.

\section{Conclusion}

We presented the first framework for zero-shot open-vocabulary detection in thermal imagery, removing the need for thermal annotations. The approach combines: large-scale synthetic supervision, a Thermal-Text Alignment Head (TTAH), and RGB-to-thermal knowledge distillation. Synthetic data generated by F-ViTA conversion of GroundingCap-1M provides diverse grounding and caption pairs for semantic learning. The distillation module transfers open-vocabulary knowledge from a frozen RGB teacher using paired RGB–thermal data, enabling effective cross-modal adaptation. Together, these components achieve strong generalization to unseen categories and challenging illumination conditions. Experiments on several thermal benchmarks confirm consistent gains over RGB-based open-vocabulary detectors, demonstrating that combining synthetic supervision with semantic alignment and distillation effectively bridges the gap between visible and thermal perception. 
% We hope this work inspires further progress toward scalable, annotation-free thermal understanding.

\section*{Acknowledgments}
This material is based upon work supported by the ASA(ALT) SBIR CCoE under Contract No. W51701-25-C-0032. Any opinions, findings and conclusions or recommendations expressed in this material are those of the author(s) and do not necessarily reflect the views of the ASA(ALT) SBIR CCoE.
\clearpage
{
    \small
    \bibliographystyle{ieeenat_fullname}
    \bibliography{main}
}

\clearpage
\setcounter{page}{1}
\maketitlesupplementary

% ===========================
% Supplementary Numbering Setup
% ===========================

% Reset counters
\setcounter{table}{0}
\setcounter{figure}{0}
\setcounter{section}{0}

% Number as Table A1, A2, A3...
\renewcommand{\thetable}{S\arabic{table}}

% Number as Figure A1, A2, A3...
\renewcommand{\thefigure}{S\arabic{figure}}

% Number as Figure A1, A2, A3...
\renewcommand{\thesection}{S\arabic{section}}

% \section{Rationale}
% \label{sec:rationale}
% % 
% Having the supplementary compiled together with the main paper means that:
% % 
% \begin{itemize}
% \item The supplementary can back-reference sections of the main paper, for example, we can refer to \cref{sec:intro};
% \item The main paper can forward reference sub-sections within the supplementary explicitly (e.g. referring to a particular experiment); 
% \item When submitted to arXiv, the supplementary will already included at the end of the paper.
% \end{itemize}
% % 
% To split the supplementary pages from the main paper, you can use \href{https://support.apple.com/en-ca/guide/preview/prvw11793/mac#:~:text=Delete%20a%20page%20from%20a,or%20choose%20Edit%20%3E%20Delete).}{Preview (on macOS)}, \href{https://www.adobe.com/acrobat/how-to/delete-pages-from-pdf.html#:~:text=Choose%20%E2%80%9CTools%E2%80%9D%20%3E%20%E2%80%9COrganize,or%20pages%20from%20the%20file.}{Adobe Acrobat} (on all OSs), as well as \href{https://superuser.com/questions/517986/is-it-possible-to-delete-some-pages-of-a-pdf-document}{command line tools}.

\section{Implementation Details}
\label{sec:suppl_impl}

We build our framework on Grounding DINO, extending it with Thermal Adapters and TTAH for thermal domain adaptation. All experiments use MMDetection v3.2 with mixed precision and gradient checkpointing. A pretrained Grounding DINO (Swin-T) serves as the frozen RGB teacher, while the thermal student is initialized from the same weights and trained on (1) synthetic thermal data from F-ViTA and (2) paired RGB–thermal datasets (MMPD, Multi-Spectral Stereo, M\textsuperscript{3}FD). Training uses roughly 250K paired thermal images. Input images are resized to $800\times800$. The detector processes \texttt{p4} and \texttt{p5} encoder features (resized to $27\times27$ and $20\times20$) concatenated into one token sequence. The CLIP text encoder is frozen, while the adapters, detection head, and TTAH are optimized jointly with detection, distillation, and alignment losses. Training runs on $8\times$ RTX A6000 GPUs with a total batch size of 16 for 150K iterations. We use AdamW with a $2\times10^{-4}$ learning rate, 0.05 weight decay, cosine scheduling, a 2K warm-up, and mixed KD:Synthetic batches at a 3:2 ratio. All baseline methods are evaluated using their official pretrained weights under a shared zero-shot thermal inference protocol, with no fine-tuning applied; applying the synthetic pretraining or distillation pipeline to these baselines would alter their methods and conflate comparisons.

\begin{table}[ht!]
\centering
\caption{Zero-shot detection results on thermal datasets SMOD, MFAD, and LLVIP.}
\setlength{\tabcolsep}{3.3pt}
\renewcommand{\arraystretch}{1.0}
\begin{tabular}{@{}lcc|cc|cc@{}}
\toprule
\multirow{2}{*}{Method} &
\multicolumn{2}{c|}{SMOD} &
\multicolumn{2}{c|}{MFAD} &
\multicolumn{2}{c}{LLVIP} \\
\cmidrule(lr){2-7}
 & AP & AP$_{50}$ & AP & AP$_{50}$ & AP & AP$_{50}$ \\
\midrule
G-DINO & 0.135 & 0.226 & 0.093 & 0.169 & 0.522 & 0.746 \\
MM-GDINO & 0.140 & 0.237 & 0.099 & 0.177 & 0.561 & 0.800 \\
LLMDet & 0.133 & 0.229 & 0.102 & 0.182 & 0.517 & 0.745 \\
\textbf{Ours} & \textbf{0.152} & \textbf{0.267} & \textbf{0.100} & \textbf{0.180} & \textbf{0.566} & \textbf{0.856} \\
\rowcolor{lightgray}
G-DINO (FT) & 0.274 & 0.357 & 0.194 & 0.354 & 0.706 & 0.979 \\
\bottomrule
\end{tabular}
\label{tab:smod_mfad_llvip}
\end{table}

\section{Additional detection results}

In this section, we provide extended quantitative results to complement those presented in the main paper. We first report additional zero-shot detection results on the SMOD, MFAD, and LLVIP datasets, which were not included in the main manuscript. As shown in Table~\ref{tab:smod_mfad_llvip}, our method consistently outperforms existing RGB open-vocabulary detectors across all three benchmarks and evaluation metrics, demonstrating its robustness and generalization ability in diverse thermal domains.

\cref{tab:smod_mfad_llvip} and \cref{tab:thermal metrics} also include the fully fine-tuned results for G-DINO (FT) across all thermal datasets evaluated in this work, covering FLIR-Aligned, FLIR-V2, CAMEL, Utokyo, SMOD, MFAD, and LLVIP. For completeness, we list alongside them the corresponding zero-shot performance of the RGB-trained open-vocabulary models evaluated in the main paper, specifically G-DINO, MM-GDINO, and LLMDet. Zero-shot results for GLIP, T-Rex2, and YOLO-World were already presented in the main manuscript and are therefore omitted here.

Comparing the G-DINO (FT) results with the zero-shot baselines highlights the level of improvement achievable when supervised thermal annotations are available. Across all datasets, fine-tuned G-DINO achieves significantly higher performance, reflecting the benefits of domain-specific training. At the same time, the difference between zero-shot G-DINO and G-DINO (FT) provides a useful reference for understanding the difficulty of transferring RGB models to the thermal domain without supervision.

Our method is designed to reduce this gap using synthetic thermal supervision and RGB to thermal cross-modal distillation rather than manual labeling. The combined comparisons across all seven datasets present a comprehensive picture of both zero-shot and fully fine-tuned performance and show how closely our approach approaches the supervised upper bound while remaining annotation-free.

\begin{table*}[ht!]
\centering
\caption{Zero-shot detection results on thermal datasets FLIR-Aligned, FLIR-V2, CAMEL, and Utokyo.}
\setlength{\tabcolsep}{4.7pt}
\renewcommand{\arraystretch}{1.0}
\begin{tabular}{lccc|ccc|ccc|ccc}
\toprule
\multirow{2}{*}{Method} &
\multicolumn{3}{c|}{FLIR-Aligned} &
\multicolumn{3}{c|}{FLIR-V2} &
\multicolumn{3}{c|}{CAMEL} &
\multicolumn{3}{c}{Utokyo} \\
\cmidrule(lr){2-13}
 & AP & AP$_{50}$ & AP$_{75}$ & AP & AP$_{50}$ & AP$_{75}$ & AP & AP$_{50}$ & AP$_{75}$ & AP & AP$_{50}$ & AP$_{75}$ \\
\midrule
GLIP & 0.251 & 0.471 & 0.226 & 0.025 & 0.041 & 0.028 & 0.186 & 0.324 & 0.242 & 0.049 & 0.093 & 0.046 \\
T-Rex2 & 0.276 & 0.514 & 0.255 & 0.033 & 0.048 & 0.036 & 0.213 & 0.345 & 0.274 & 0.049 & 0.101 & 0.043 \\
YOLO-World & 0.266 & 0.487 & 0.241 & 0.029 & 0.044 & 0.032 & 0.197 & 0.336 & 0.256 & 0.050 & 0.101 & 0.043 \\
G-DINO & 0.337 & \underline{0.636} & 0.313 & \underline{0.081} & \underline{0.144} & \underline{0.078} & \underline{0.482} & \underline{0.729} & \textbf{0.543} & \underline{0.050} & 0.093 & 0.169 \\
MM-GDINO & 0.354 & 0.619 & \textbf{0.371} & 0.042 & 0.060 & 0.048 & 0.273 & 0.369 & 0.340 & 0.047 & 0.095 & 0.042 \\
LLMDet & \underline{0.359} & 0.628 & 0.348 & 0.048 & 0.075 & 0.051 & 0.383 & 0.560 & 0.439 & 0.050 & 0.102 & 0.045 \\
\textbf{Ours} & \textbf{0.372} & \textbf{0.664} & \underline{0.359} & \textbf{0.096} & \textbf{0.173} & \textbf{0.091} & \textbf{0.511} & \textbf{0.758} & \underline{0.525} & \textbf{0.065} & \textbf{0.137} & \textbf{0.054} \\
% \vspace{-3mm}\\
\rowcolor{lightgray}
G-DINO (FT) & 0.414 & 0.749 & 0.414 & 0.200 & 0.345 & 0.203 & 0.478 & 0.684 & 0.533 & 0.129 & 0.252 & 0.115 \\
\bottomrule
\end{tabular}
\label{tab:thermal metrics}
\end{table*}

\section{Related works}
\label{sec: related works}

\noindent {\bf{Open-vocabulary object detection.}}
Open-vocabulary object detection (OVD) enables recognition beyond fixed categories through large-scale vision–language pre-training. Early works such as GLIP \cite{li2022grounded} and Grounding DINO \cite{liu2024grounding} aligned visual regions with text to achieve open-set detection. Recent methods, including YOLO-World \cite{cheng2024yolo}, T-Rex2 \cite{jiang2024t}, and LLMDet \cite{fu2025llmdet}, improved efficiency and semantic grounding using one-stage architectures, prompt synergy, and large language model supervision. However, these advances remain confined to the RGB domain, and extending them to thermal imagery demands new strategies for cross-modal alignment and domain adaptation.\\
%\subsection{Synthetic thermal data and domain adaptation}
\noindent {\bf{Synthetic thermal data and domain adaptation.}}
Thermal object detection suffers from limited labeled data, motivating synthetic generation and domain adaptation to transfer knowledge from RGB imagery. Early studies such as SSTN \cite{munir2021sstn} and Domain-Adaptive Pedestrian Detection \cite{guo2019domain} used self-supervised and adversarial methods to reduce modality gaps, while ThermalSynth \cite{madan2023thermalsynth} and Hamrell and Karlholm \cite{hamrell2021image} improved realism via simulation and GAN-based translation. Meta-UDA \cite{vs2022meta} introduced meta-learning for unsupervised adaptation across environments. More recent foundation-guided models, including F-ViTA \cite{paranjape2025f} and ThermalGen \cite{xiao2025thermalgen}, achieve semantically consistent visible-to-thermal synthesis, marking a shift toward scalable cross-modal generation. Yet, most approaches treat thermal adaptation and semantic grounding separately, and our work unifies them through synthetic supervision, language alignment, and cross-modal distillation for zero-shot thermal detection.
Together, advances in open-vocabulary detection and synthetic thermal generation lay the groundwork for scalable perception beyond labeled data. Building on these trends, our work unifies language supervision, synthetic thermal pre-training, and cross-modal alignment to achieve zero-shot thermal object detection.

\section{Pre-training datasets and backbones}

To contextualize the capabilities of the baseline open-vocabulary detectors evaluated in this work, Table~\ref{tab:backbone_pretrain} summarizes their backbone architectures and pre-training corpora. Most RGB open-vocabulary models, including GLIP, Grounding-DINO, MM-GDINO, and LLMDet, rely on Swin-T backbones trained on large-scale grounding datasets such as Objects365, GoldG, and various captioning or region-level corpora. YOLO-World instead adopts a YOLOv8-L backbone combined with mixture-of-caption datasets to enhance efficiency and prompt alignment, while T-Rex2 is trained on a diverse 10M-image collection aggregated from multiple sources. These pre-training differences influence the semantic coverage, grounding quality, and transferability of each model, and highlight that all baselines are optimized primarily for RGB imagery. This underscores the challenge of applying them directly to thermal data and motivates our approach for bridging the modality gap through synthetic thermal supervision and cross-modal distillation.

\begin{table}[ht!]
\centering
\caption{Backbone architectures and pre-training datasets used by baseline open-vocabulary detectors evaluated in this work.}
\setlength{\tabcolsep}{3pt}
\renewcommand{\arraystretch}{1.05}
\begin{tabular}{@{}lcc@{}}
\toprule
Method & Backbone & Pre-training data \\
\midrule
GLIP \cite{li2022grounded} & Swin & O365, GoldG, Cap4M \\
T-Rex2 \cite{jiang2024t} & Swin & \small{10M data from various resources} \\
YOLO-World \cite{cheng2024yolo} & YOLO & O365, GoldG, CC3M \\
G-DINO \cite{liu2024grounding} & Swin & O365, GoldG, Cap4M \\
MM-GDINO  & Swin & O365, GoldG, GRIT, V3Det \\
LLMDet \cite{fu2025llmdet} & Swin & GroundingCap-1M \\
\multirow{2}{*}{Ours}  & \multirow{2}{*}{Swin} & Synthetic GroundingCap-1M \\
 & & M\textsuperscript{3}FD, MMPD, MS\textsuperscript{2}\\
\bottomrule
\end{tabular}
\label{tab:backbone_pretrain}
\end{table}

\section{TTAH Drift Regularization Sensitivity}

We analyze the sensitivity of TTAH to the drift regularization weight $\lambda_{\text{drift}}$ by varying it over the range $\{0, 0.25, 1, 5, 10\}$ on FLIR-Aligned. Aside from $\lambda_{\text{drift}}$, all other loss terms are normalized to comparable value ranges. The results show stable behavior across this range: the model peaks at $\lambda_{\text{drift}}=0.25$ (AP=0.259) and degrades gradually for larger values (AP=0.247 at $\lambda_{\text{drift}}=1$, AP=0.226 at $\lambda_{\text{drift}}=5$), indicating robustness rather than sensitivity to this hyperparameter. Setting $\lambda_{\text{drift}}=0$ removes the drift constraint entirely, allowing unconstrained adaptation that hurts generalization. These runs were conducted on a single GPU due to resource constraints.

\section{Synthetic Data Domain Shift Analysis}

To characterize the domain shift between our synthetic thermal data and real thermal benchmarks, we compute FID scores between the synthetic dataset generated via F-ViTA and the FLIR-Aligned validation set. The FID between our synthetic dataset and FLIR-Aligned is 41.40, comparable to the FID between two real thermal benchmarks, FLIR-V2 and FLIR-Aligned (38.74), indicating a similar domain shift magnitude. Caption adaptation removes RGB-specific descriptors only—such as color terms and lighting conditions—while preserving object identity and modality-invariant spatial relations.

\begin{table*}[t]
\centering
\setlength{\tabcolsep}{10pt}
\renewcommand{\arraystretch}{1.2}

\begin{tabular}{c c}

    % ---------------- Row 1 ----------------
    \begin{minipage}[t]{0.45\linewidth}\vspace{0pt}
        \centering
        \includegraphics[width=.75\linewidth]{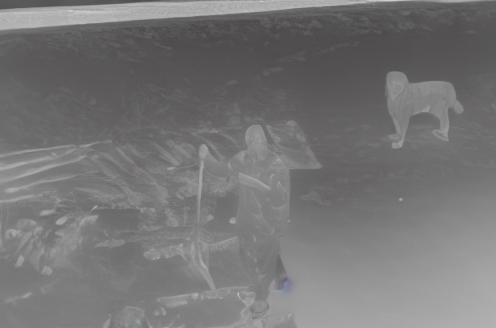}
    \end{minipage}
    &
    \begin{minipage}[t]{0.45\linewidth}\vspace{0pt}
        \small
        A person stands in a snowy outdoor area, holding a cold, shovel-shaped object whose
        metal blade appears significantly cooler than the surroundings. The individual’s warm
        head and torso contrast with the insulated, cooler areas of their clothing. Beside them,
        a dog displays a distinct warm body signature with a cooler collar visible around its
        neck. The ground shows cold snow with slightly warmer linear patterns indicating tire
        tracks, while the background includes a cold, snow-covered paved surface and
        snow-covered vegetation on the left.
    \end{minipage}
    \\[1.2em]

    % ---------------- Row 2 ----------------
    \begin{minipage}[t]{0.45\linewidth}\vspace{0pt}
        \centering
        \includegraphics[width=.75\linewidth]{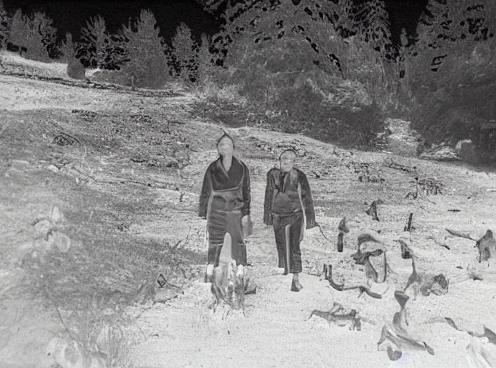}
    \end{minipage}
    &
    \begin{minipage}[t]{0.45\linewidth}\vspace{0pt}
        \small
        Two warm human figures walk across a cool, grassy hillside, their bodies showing
        strong heat signatures from the head and torso with noticeably cooler patterns across
        their layered clothing. The individual on the left carries a cooler, bag-shaped object
        whose low thermal emission contrasts with the warmer hand holding it. The terrain
        beneath them appears mostly cool, with mixed patches of slightly warmer vegetation
        and scattered cooler leaf material. In the background, multiple trees appear as large,
        mostly cool vertical structures, including evergreens with very low heat emission and
        deciduous trees showing faint, uneven warmth. The overall scene indicates a cool
        outdoor environment with soft thermal contrast across the landscape.
    \end{minipage}
    \\[1.2em]

    % ---------------- Row 3 ----------------
    \begin{minipage}[t]{0.45\linewidth}\vspace{0pt}
        \centering
        \includegraphics[width=.75\linewidth]{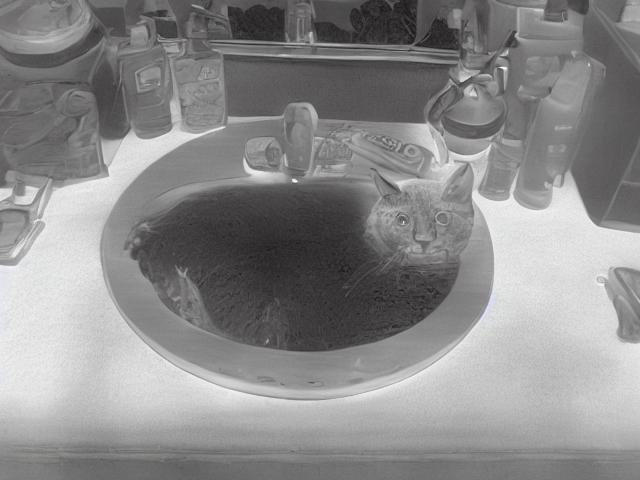}
    \end{minipage}
    &
    \begin{minipage}[t]{0.45\linewidth}\vspace{0pt}
        \small
        A warm cat sits inside a cool, bowl-shaped sink basin, its body showing a concentrated
        heat signature with the head and torso emitting the strongest thermal contrast. The sink
        surrounding the cat appears uniformly cool, with the metal faucet registering as even
        colder. Spread around the countertop are numerous toiletry items displaying varying
        thermal profiles—most of them cool, with a few showing mild warmth likely due to
        recent handling or sun exposure. The countertop itself emits a broad, stable cool
        temperature, while the mirror behind it reflects minimal thermal information. The cat’s
        warm body sharply contrasts the cool surfaces around it, making it the dominant
        thermal feature of the scene.
    \end{minipage}
    \\[1.2em]

    % ---------------- Row 4 ----------------
    \begin{minipage}[t]{0.45\linewidth}\vspace{0pt}
        \centering
        \includegraphics[width=.35\linewidth]{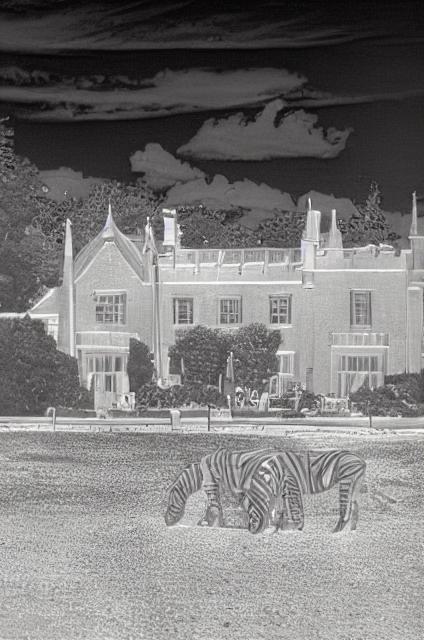}
    \end{minipage}
    &
    \begin{minipage}[t]{0.45\linewidth}\vspace{0pt}
        \small
        Two large, warm zebra-shaped heat signatures stand on a cool grassy field, their
        bodies emitting consistent warmth with slightly cooler patterns along the legs and
        extremities. The ground beneath them appears broadly cool with subtle variations
        caused by grass density and soil composition. Behind the animals, a large stone
        mansion rises as a predominantly cool structure, its walls and roof retaining very
        little heat and displaying uniform low-temperature surfaces. The chimneys and
        architectural details show the same cold thermal characteristics. Above the building,
        the sky registers as a broad, cold background with no significant thermal emission.
        The warm zebras create a striking contrast against the cool grass and the colder
        stone architecture behind them.
    \end{minipage}

\end{tabular}

\captionof{figure}{Qualitative thermal examples with descriptions.}
\label{fig:thermal_qualitative}
\end{table*}

Functionally, a detector trained exclusively on synthetic data (without any subsequent distillation on real paired data) achieves 0.372 mAP on FLIR-Aligned, with near parity on the \emph{person} class relative to fully fine-tuned G-DINO. This confirms that the synthetic initialization provides meaningful semantic coverage and serves as an effective starting point for cross-modal distillation, with remaining mismatches corrected through subsequent training on real RGB--thermal pairs.

\section{Limitations}
Our approach has several limitations. First, the synthetic thermal data used during pre-training does not fully reproduce sensor characteristics such as emissivity variations, temperature gradients, and noise patterns, leaving a residual domain gap between synthetic and real thermal imagery. Second, the cross-modal distillation stage depends on the accuracy of the RGB teacher detector, and any errors in the teacher's predictions can be transferred to the thermal student and limit zero-shot performance. A further limitation arises from the textual supervision used during pre-training. Because the synthetic dataset is constructed from region-level grounding phrases, the model receives many more region-level annotations than detailed image-level descriptions. This imbalance can cause the language-conditioned components to generate shorter or less informative captions, even when detailed descriptions are requested. Incorporating more descriptive region-level text or stronger image-level supervision could help address this limitation and further improve semantic grounding.

\subsection{Visualizations of synthetic training data}
\label{sec:thermal_caption_visualization}

In \cref{fig:thermal_qualitative}, we illustrate several examples of the synthetic thermal images and their corresponding textual annotations used during pre-training. These samples are drawn from our synthetic dataset, where thermal images are produced through RGB-to-thermal translation and paired with region-level grounding phrases derived from the original RGB captions. The annotations accurately identify the main objects within each scene, demonstrating that the grounding-based text supervision provides reliable category signals for open-vocabulary thermal detection.

\end{document}